\documentclass[dvipsnames,format=sigconf,anonymous=false,review=false]{acmart}

\makeatletter
\def\@ACM@checkaffil{
    \if@ACM@instpresent\else
    \ClassWarningNoLine{\@classname}{No institution present for an affiliation}%
    \fi
    \if@ACM@citypresent\else
    \ClassWarningNoLine{\@classname}{No city present for an affiliation}%
    \fi
    \if@ACM@countrypresent\else
        \ClassWarningNoLine{\@classname}{No country present for an affiliation}%
    \fi
}
\makeatother

\usepackage{geometry, graphicx}
\usepackage{colortbl}
\usepackage{tikz}
\usepackage{hyperref}
\usepackage{color}
\usepackage{comment}
\usepackage{multirow}
\usepackage{relsize}
\usepackage{subcaption}
\definecolor{lightgray}{gray}{0.9}

\AtBeginDocument{%
  \providecommand\BibTeX{{%
    \normalfont B\kern-0.5em{\scshape i\kern-0.25em b}\kern-0.8em\TeX}}}

\setcopyright{acmcopyright}
\copyrightyear{2018}
\acmYear{2018}
\acmDOI{10.1145/1122445.1122456}

\copyrightyear{2023}
\acmYear{2023}
\setcopyright{acmlicensed}\acmConference[GECCO '23 Companion]{Genetic and Evolutionary Computation Conference Companion}{July 15--19, 2023}{Lisbon, Portugal}
\acmBooktitle{Genetic and Evolutionary Computation Conference Companion (GECCO '23 Companion), July 15--19, 2023, Lisbon, Portugal}
\acmPrice{15.00}
\acmDOI{10.1145/3583133.3596336}
\acmISBN{979-8-4007-0120-7/23/07}

\usepackage{algorithm}
\usepackage{algpseudocode}
\usepackage{diagbox}
\usepackage{float}

\algrenewcommand\algorithmicrequire{\textbf{Data:}}
\algrenewcommand\algorithmicensure{\textbf{Result:}}


\begin{document}


\title{Can the Problem-Solving Benefits of Quality Diversity Be Obtained Without Explicit Diversity Maintenance?}

\author{Ryan Boldi}
\affiliation{%
  \institution{University of Massachusetts Amherst}
  \city{Amherst}
  \state{MA}
}
\email{rbahlousbold@umass.edu}

\author{Lee Spector}
\affiliation{%
  \institution{Amherst College}
  \city{Amherst}
  \state{MA}
}
\email{lspector@amherst.edu}

\begin{abstract}
When using Quality Diversity (QD) optimization to solve hard exploration or deceptive search problems, we assume that diversity is extrinsically valuable. This means that diversity is important to help us reach an objective, but is not an objective in itself. Often, in these domains, practitioners benchmark their QD algorithms against single objective optimization frameworks. In this paper, we argue that the correct comparison should be made to \emph{multi-objective} optimization frameworks. This is because single objective optimization frameworks rely on the aggregation of sub-objectives, which could result in decreased information that is crucial for maintaining diverse populations automatically. In order to facilitate a fair comparison between quality diversity and multi-objective optimization, we present a method that utilizes dimensionality reduction to automatically determine a set of behavioral descriptors for an individual, as well as a set of objectives for an individual to solve. Using the former, one can generate solutions using standard quality diversity optimization techniques, and using the latter, one can generate solutions using standard multi-objective optimization techniques. This allows for a level comparison between these two classes of algorithms, without requiring domain and algorithm specific modifications to facilitate a comparison.
\end{abstract}



\maketitle

\section{Introduction and Background}
Quality Diversity (QD) optimization is a relatively recent advancement in the evolutionary computation literature where a diverse set of high performing individuals are maintained over the course of the run. This results in \emph{divergent}, rather than the traditional \emph{convergent} evolutionary search \citep{pugh_quality_2016}.

There are often two reasons to apply quality diversity optimization. The first of them is to generate a large archive of qualitatively diverse individuals that solve certain problems. For example, finding diverse sets of robot behaviors \citep{cully2015evolving, kim2017learning} or creating diverse video game or training scenarios \cite{fontaine_evaluating_2022, earle2022videogame}. The second reason is generally to solve hard exploration problems that often have deceptive reward signals. For example, \citet{lehman_abandoning_2011} explore using Novelty Search (NS) to solve a deceptive maze, where there is a single goal, although many ways to solve this goal. They also used NS to evolve a controller for bipedal locomotion that outperformed fitness-based search with the particular fitness function and selection scheme studied. Some more recent examples are the QD-Maze brought forward by \citet{pugh_confronting_2015}, or the modular robotics domain used by \citet{nordmoen2021map}, which also have a single (real) objective. Despite the only goal for experimenters being solving the single objective, there is the assumption that diversity here is instrumental to solving the task. The intuition behind this makes sense: diverse low-performing solutions might be stepping stones that lead to high performing solutions in the future. This paper focuses on bench-marking Quality Diversity on the second use case, as an exploration algorithm that ultimately solves a single problem, although possibly in unexpected ways.

Multi-objective optimization (MOO) is a common paradigm in optimization literature that attempts to optimize for multiple objectives at the same time. For example, \emph{non dominated sorting genetic algorithm} (NSGA2) \citep{deb2002nsga2} and \emph{strength pareto evolutionary algorithm} (SPEA2) \citep{zitzler_spea2_2001} both deal with situations where there are multiple different objectives that often have numerous trade-offs for each other. Lexicase selection, a parent selection technique, has also been found to be useful at optimizing for multiple objectives \citep{la2019probabilistic}. There are often many different solutions that can be made by having different trade-offs between objectives, which can be visualized as existing on a Pareto front. These solutions could be highly diverse, as individuals that trade off between objectives differently would likely have qualitatively different behavior \citep{riolo_lexicase_2016, Laumanns2002ArchivingWG} These methods can therefore be used as a form of implicit quality diversity optimization: the quality comes from solving objectives, and the diversity comes from solving different combinations of objectives.
\newsavebox{\largestimage}
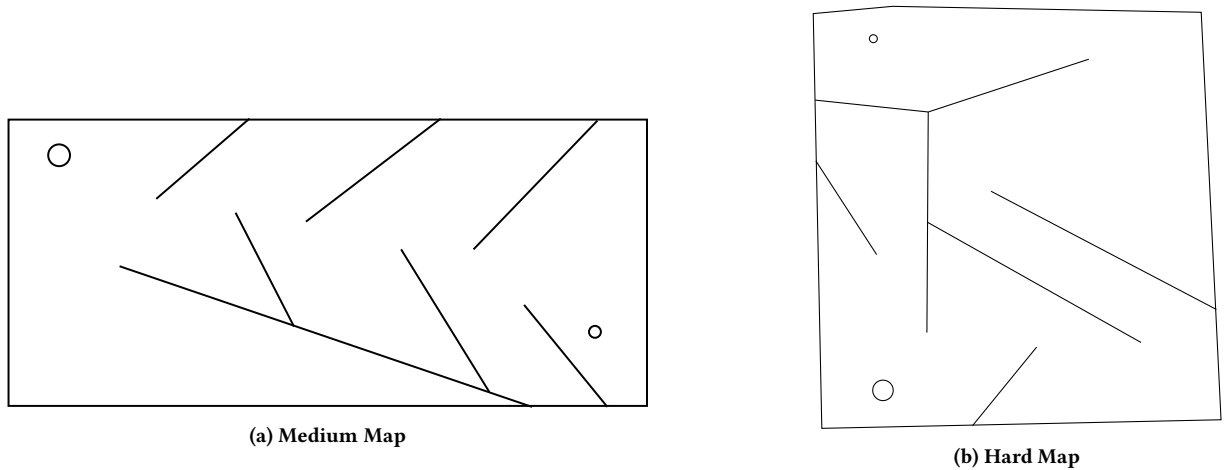
\begin{figure*}[t]
     \centering
     \begin{subfigure}[b]{0.55\textwidth}
         \centering
\tikzset{every picture/.style={line width=0.75pt}} 
\begin{tikzpicture}[x=0.75pt,y=0.75pt,yscale=-1.5,xscale=1.5]

\draw   (110.33,99.33) -- (325,99.33) -- (325,195.67) -- (110.33,195.67) -- cycle ;
\draw   (123.67,111.33) .. controls (123.67,109.31) and (125.31,107.67) .. (127.33,107.67) .. controls (129.36,107.67) and (131,109.31) .. (131,111.33) .. controls (131,113.36) and (129.36,115) .. (127.33,115) .. controls (125.31,115) and (123.67,113.36) .. (123.67,111.33) -- cycle ;
\draw    (191.33,99) -- (160,126) ;
\draw    (147.67,148.67) -- (286.33,196) ;
\draw    (255.67,99) -- (210.33,133.67) ;
\draw    (308.33,99.67) -- (266.67,143) ;
\draw    (186.67,130.67) -- (206.33,169) ;
\draw    (242.33,143) -- (272,191) ;
\draw    (283.67,161.67) -- (311.67,196) ;
\draw   (305.5,170.75) .. controls (305.5,169.65) and (306.4,168.75) .. (307.5,168.75) .. controls (308.6,168.75) and (309.5,169.65) .. (309.5,170.75) .. controls (309.5,171.85) and (308.6,172.75) .. (307.5,172.75) .. controls (306.4,172.75) and (305.5,171.85) .. (305.5,170.75) -- cycle ;
\end{tikzpicture}
         \caption{Medium Map}
         \label{fig:MediumMap}\hspace*{-2.9em}
     \end{subfigure}
     \hfill
     \savebox{\largestimage}{
\begin{tikzpicture}[x=0.75pt,y=0.75pt,yscale=-1,xscale=1]

\draw   (299.67,227) -- (309.67,432.67) -- (108.33,437) -- (104,227.67) -- (144,224) -- cycle ;
\draw    (105,271.33) -- (162,277.33) ;
\draw    (162,277.33) -- (243,250.67) ;
\draw    (162,277.33) -- (161.33,388.67) ;
\draw    (105.33,302) -- (136,349.33) ;
\draw    (161.67,333) -- (269.33,393.67) ;
\draw    (193.67,317.33) -- (307.33,377) ;
\draw    (216.67,396) -- (184.33,435.67) ;
\draw   (134,417.83) .. controls (134,414.98) and (136.31,412.67) .. (139.17,412.67) .. controls (142.02,412.67) and (144.33,414.98) .. (144.33,417.83) .. controls (144.33,420.69) and (142.02,423) .. (139.17,423) .. controls (136.31,423) and (134,420.69) .. (134,417.83) -- cycle ;
\draw   (132.33,240.33) .. controls (132.33,239.23) and (133.23,238.33) .. (134.33,238.33) .. controls (135.44,238.33) and (136.33,239.23) .. (136.33,240.33) .. controls (136.33,241.44) and (135.44,242.33) .. (134.33,242.33) .. controls (133.23,242.33) and (132.33,241.44) .. (132.33,240.33) -- cycle ;
\end{tikzpicture}
}
     \begin{subfigure}[b]{0.42\textwidth}
         \centering
         \usebox{\largestimage}
         \caption{Hard Map}
         \label{fig:hardmaze}
         
     \end{subfigure}
        \caption{Two deceptive mazes, adapted from \citep{lehman_abandoning_2011}. In both maps, the large circle is the starting point of the robot and the small circle represents the goal. Both maps are deceptive in the sense that the path that minimizes euclidean distance to the goal will get stuck at local optima.}
        \label{fig:mazes}
\end{figure*}

How do you determine whether a quality diversity algorithm is performing well in hard exploration or deceptive domains? Previous studies compare quality diversity to using single objective optimization in reaching the objective. For example, \citet{lehman_abandoning_2011} compare novelty search to single-objective fitness based optimization in maze domains. \citet{pugh_quality_2016} compare quality diversity techniques to using only a single distance metric as an objective. \citet{gaier_are_2019} compare using MAP-Elites to a (single) image similarity objective in evolving a target image. There are notably few studies comparing it to multi-objective or multi-modal optimization. \citet{vassiliades_comparing_2017} compared MAP-Elites and NS to other multi-modal selection schemes for a maze navigation task. \citet{nordmoen2021map} compared MAP-Elites \citep{mouret_illuminating_2015} to a single- and multi-objective optimization algorithm. However, the multi-objective algorithms simply use diversity as a secondary objective, which does not necessarily provide a fair ground between QD algorithms and MOOs as QD has access to all dimensions of diversity, where MOOs simply get access to an aggregated version of this information. Other work looked at combining QD with MOO, but not on comparing them to each other (partly due to them having different goals in the majority of the cases) \citep{pierrot2022multi, mouret2020quality}.

When comparing QD's problem solving ability to single objective or limited multi-objective optimization paradigms, the information accessible to both systems is not the same, preventing a fair comparison. In this paper, we consider the problem of providing quality diversity algorithms with the same information as that available to multi-objective optimization. This allows for both MOOs and QD to be compared faithfully in their ability to maximize objective(s) where diversity is instrumental (extrinsic), as opposed to being a goal that we are trying to optimize in and of itself (intrinsic).

To do this, we propose the use of a dimensionality reduction technique to generate the behavioral descriptors for the quality diversity methods, and a slightly augmented version of this model to generate the objective values for MOO. This allows for a faithful comparison between both techniques as they will have access to the same performance signal.



\section{Motivation}

To motivate the use of comparison scheme like this, we discuss the potential effects of learning the fitness function from hand-written measures (as it usually is in the literature), and provide a motivating example that shows how optimizing for combinations of these objectives through multi-objective optimization could result in the maintenance of diversity \emph{automatically}.

A common motivating example for quality diversity algorithms aimed at solving deceptive single objective problems is one of the deceptive hard maze. Two mazes that are of medium and hard difficulty can be found in Figure~\ref{fig:mazes}. These mazes are both deceptive as an individual greedily moving towards the goal will get stuck at sub-optimal traps. For this reason, diversity must be emphasized to ensure that the goal is actually reached. This fits into scope of the problems discussed in this paper as the sole objective is to reach the goal, yet diversity is intstrumental in reaching this objective.

Consider what an arbitrary QD algorithm could do in this scenario. For example, MAP-elites \citep{mouret_illuminating_2015} could be implemented with hand written behavioral descriptors such as 1) distance to goal on $x$ axis, 2) distance to goal on $y$ axis, 3) total distance travelled, and perhaps 4) number of turns taken. Using measure functions like these, solving deceptive mazes like above would be relatively straightforward, as demonstrated by \citet{pugh_quality_2016}.

\paragraph{Why not single objectives?}

Single objectives often rely on an aggregation of sub-objectives that each represent qualitatively different goals. This aggregation procedure results in a loss of information regarding the performance of an individual. 

Instead, we use multi-objective optimization directly on the sub-objectives to help facilitate the discovery of high performing individuals. With MOOs, then, finding solutions that optimize combinations of the sub-objectives creates diversity in the population similar to that with QD. However, the diversity that results from this is implicitly optimized for, as opposed to it being explicitly optimized for like in QD. What this means is QD maintains diversity as an objective in-and-of itself, whereas MOOs only optimize for their given objectives.

In order to use multi-objective optimization on this domain using the scheme we present in this paper, we make an assumption that the fitness function (proximity to goal) can be approximated as a linear combination of these 4 measures. It is important to note that the raw value of the fitness function need not matter, as long as it matches the original fitness function in ranking the individuals. It is clear that both $x$ and $y$ distance are negatively correlated with proximity to the goal. For the sake of the example, let us suppose that the total distance travelled could also have a negative (yet smaller in magnitude) correlation to proximity to the goal. Finally, let us say that the number of turns is neutral to fitness (individuals that turn more on average perform no better than those that turn less).

We then set each feature as its own objective for multi-objective optimization. Each individual is evaluated on all 4 of these metrics, and individuals are selected based on the multi-objective optimization algorithm being used. Consider an individual that has perfectly matched the $x$ location of the goal. This individual will likely be selected regardless of its $y$ location, distance, or turn values. Consider an individual that is very far from the goal, yet has covered a large amount of distance. This individual would also be selected, regardless of its distance to the goal. These different trade offs result in a diversity of individuals that solve different combinations of the sub-problems. When using a multi-objective selection scheme like Lexicase selection, this diversity is maintained automatically until convergence to a final goal. \citep{helmuth_effects_2016}

Importantly, these systems have access to the same information as the quality diversity optimization algorithms. However, the key difference here is that MOOs could solve these deceptive problems \emph{without explicitly maintaining diversity}. If QD algorithms can outperform MOOs in this domain, this would mean that the diversity that that specific QD algorithm maintains is instrumental in overcoming the deception in this domain. In the next section, we discuss how to extend this comparison scheme to any domain, regardless of the existence of human-written behavioral descriptors.

\section{Comparison Scheme}
In this section, we will bring together all the ideas presented in this paper into a simple scheme to fairly compare quality diversity algorithms with objective-based algorithms. We operate under the assumption that humans have little intuition about the domain, and that reasonable choices behavioral descriptors or objectives will therefore not be obvious in advance.

\subsection{From Phenotype to Measures}

Consider learning of a set of measures $m(\theta)$ as an unsupervised learning task. This can be learned with a variational autoencoder (VAE, \citep{doersch2016tutorial}) as done in some previous work in QD \citep{cully_autonomous_2019, grillotti_unsupervised_2022}. 

\begin{figure}[t]

\tikzset{every picture/.style={line width=0.75pt}} 

\begin{tikzpicture}[x=0.75pt,y=0.75pt,yscale=-1,xscale=1]

\draw   (53,21.5) -- (146,49.4) -- (146,105.6) -- (53,133.5) -- cycle ;
\draw   (149.67,50.32) -- (167.33,50.32) -- (167.33,104.68) -- (149.67,104.68) -- cycle ;
\draw   (265,133.5) -- (172,105.6) -- (172,49.4) -- (265,21.5) -- cycle ;

\draw (71,69.5) node [anchor=north west][inner sep=0.75pt]   [align=left] {Encoder};
\draw (20,70.9) node [anchor=north west][inner sep=0.75pt]    {$p( \theta )$};
\draw (192,69.5) node [anchor=north west][inner sep=0.75pt]   [align=left] {Decoder};
\draw (268,68.9) node [anchor=north west][inner sep=0.75pt]    {$\hat{p}( \theta )$};
\draw (140,109) node [anchor=north west][inner sep=0.75pt]    {$ \begin{array}{l}
z( \theta )\\
\end{array}$};

\end{tikzpicture}
\caption{Variational autoencoder architecture used to learn a latent embedding $z(\theta)$ of a phenotype $p(\theta)$. As this is a variational autoencoder, $z(\theta)$ is sampled from a distribution that is parameterized by the output of the encoder (not pictured here for simplicity).}\label{fig:vae1}
\end{figure}
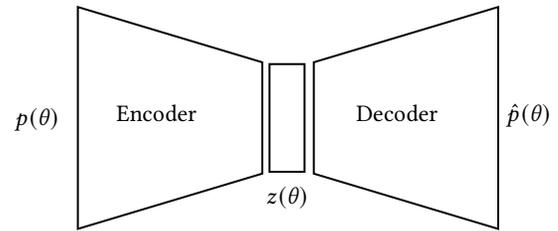

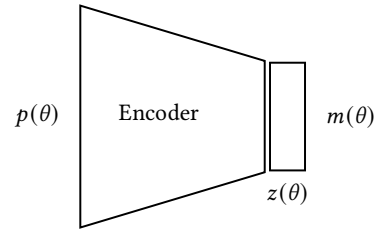
\begin{figure}[t]

\tikzset{every picture/.style={line width=0.75pt}} 

\begin{tikzpicture}[x=0.75pt,y=0.75pt,yscale=-1,xscale=1]

\draw   (80,40.9) -- (173,68.8) -- (173,125) -- (80,152.9) -- cycle ;
\draw   (175.67,69.72) -- (193.33,69.72) -- (193.33,124.08) -- (175.67,124.08) -- cycle ;

\draw (98,88.9) node [anchor=north west][inner sep=0.75pt]   [align=left] {Encoder};
\draw (46,89.3) node [anchor=north west][inner sep=0.75pt]    {$p( \theta )$};
\draw (166,127.4) node [anchor=north west][inner sep=0.75pt]    {$ \begin{array}{l}
z( \theta )\\
\end{array}$};
\draw (196,88.4) node [anchor=north west][inner sep=0.75pt]    {$ \begin{array}{l}
m( \theta )\\
\end{array}$};

\end{tikzpicture}
\caption{Using the encoder of a VAE to predict measures $m(\theta)$ of a phenotype $p(\theta)$}\label{fig:vae-measures}
\end{figure}

Figure~\ref{fig:vae1} outlines an example autoencoder architecture that could be used to learn a latent embedding of a given phenotype.
Using reconstruction loss, this autoencoder can be used to learn a compressed representation of a phenotype in lower-dimensional space. This makes it possible for QD algorithms such as MAP-Elites to cover a large set of behavioral niches without needing to store a high dimensional archive.

Figure~\ref{fig:vae-measures} shows how the architecture outline in Figure~\ref{fig:vae1} can be used to generate the qualitative measures $m({\theta})$ that can be used to emphasize diversity using one of many QD techniques.

\subsection{From Measures to Fitness}
In order to facilitate a comparison between quality diversity techniques and multi-objective optimization techniques, one should try to present them both with as close to the same information as possible. 

In order to apply multi-objective optimization, we need to extract the sub-objectives from both the phenotype $p(\theta)$ and the ground truth fitness function $f(\theta)$. This can be done by simply augmenting the dimensionality reduction architecture used to generate the measures. We assume that the ground truth fitness can be approximated by a linear combination of the measure functions. Although this is a large assumption, it is one that is commonly made in the reward learning literature \footnote{It is possible to relax this assumption to be that fitness is a non-linear function of the measures by incorporating a second multi-layer perception that learns the fitness from the output of the encoder. However, a similar de-aggregation procedure would be necessary on the penultimate layer of the newly added MLP.}. In inverse reinforcement learning (IRL), where the task is to learn a reward function that explains an expert trajectory, it is often the case that this reward function is learned as a linear combination of state features $\phi_s$ \citep{ng_200_irl}. More recent work in IRL uses a set of pretraining algorithms to learn the state features (either with an autoencoder, or predicting a forward dynamics model, etc) before learning the final linear combination of these features that results in a reward function \citep{brown_BREX_2020}.

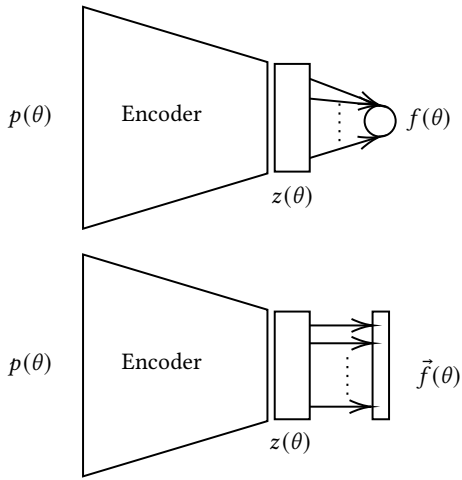
\begin{figure}

\tikzset{every picture/.style={line width=0.75pt}} 

\begin{tikzpicture}[x=0.75pt,y=0.75pt,yscale=-1,xscale=1]

\draw   (100,10) -- (193,37.9) -- (193,94.1) -- (100,122) -- cycle ;
\draw   (196.67,38.82) -- (214.33,38.82) -- (214.33,93.18) -- (196.67,93.18) -- cycle ;
\draw   (242,67.67) .. controls (242,63.34) and (245.51,59.83) .. (249.83,59.83) .. controls (254.16,59.83) and (257.67,63.34) .. (257.67,67.67) .. controls (257.67,71.99) and (254.16,75.5) .. (249.83,75.5) .. controls (245.51,75.5) and (242,71.99) .. (242,67.67) -- cycle ;
\draw    (214.33,46.33) -- (247.96,59.12) ;
\draw [shift={(249.83,59.83)}, rotate = 200.82] [color={rgb, 255:red, 0; green, 0; blue, 0 }  ][line width=0.75]    (10.93,-3.29) .. controls (6.95,-1.4) and (3.31,-0.3) .. (0,0) .. controls (3.31,0.3) and (6.95,1.4) .. (10.93,3.29)   ;
\draw    (214.33,56.33) -- (247.84,59.64) ;
\draw [shift={(249.83,59.83)}, rotate = 185.63] [color={rgb, 255:red, 0; green, 0; blue, 0 }  ][line width=0.75]    (10.93,-3.29) .. controls (6.95,-1.4) and (3.31,-0.3) .. (0,0) .. controls (3.31,0.3) and (6.95,1.4) .. (10.93,3.29)   ;
\draw    (214.33,86.33) -- (247.92,76.08) ;
\draw [shift={(249.83,75.5)}, rotate = 163.03] [color={rgb, 255:red, 0; green, 0; blue, 0 }  ][line width=0.75]    (10.93,-3.29) .. controls (6.95,-1.4) and (3.31,-0.3) .. (0,0) .. controls (3.31,0.3) and (6.95,1.4) .. (10.93,3.29)   ;
\draw  [dash pattern={on 0.84pt off 2.51pt}]  (229.33,58.83) -- (229.33,79.83) ;
\draw   (100,135) -- (193,162.9) -- (193,219.1) -- (100,247) -- cycle ;
\draw   (196.67,163.82) -- (214.33,163.82) -- (214.33,218.18) -- (196.67,218.18) -- cycle ;
\draw    (214.33,170.83) -- (244.33,170.83) ;
\draw [shift={(246.33,170.83)}, rotate = 180] [color={rgb, 255:red, 0; green, 0; blue, 0 }  ][line width=0.75]    (10.93,-3.29) .. controls (6.95,-1.4) and (3.31,-0.3) .. (0,0) .. controls (3.31,0.3) and (6.95,1.4) .. (10.93,3.29)   ;
\draw    (214.33,179.83) -- (244.33,179.83) ;
\draw [shift={(246.33,179.83)}, rotate = 180] [color={rgb, 255:red, 0; green, 0; blue, 0 }  ][line width=0.75]    (10.93,-3.29) .. controls (6.95,-1.4) and (3.31,-0.3) .. (0,0) .. controls (3.31,0.3) and (6.95,1.4) .. (10.93,3.29)   ;
\draw    (214.33,211.83) -- (243.33,211.83) ;
\draw [shift={(245.33,211.83)}, rotate = 180] [color={rgb, 255:red, 0; green, 0; blue, 0 }  ][line width=0.75]    (10.93,-3.29) .. controls (6.95,-1.4) and (3.31,-0.3) .. (0,0) .. controls (3.31,0.3) and (6.95,1.4) .. (10.93,3.29)   ;
\draw  [dash pattern={on 0.84pt off 2.51pt}]  (233.33,186.83) -- (233.33,206.5) ;
\draw   (246.02,163.82) -- (254.33,163.82) -- (254.33,218.18) -- (246.02,218.18) -- cycle ;

\draw (118,183) node [anchor=north west][inner sep=0.75pt]   [align=left] {Encoder};
\draw (61,183.4) node [anchor=north west][inner sep=0.75pt]    {$p( \theta )$};
\draw (186,221.5) node [anchor=north west][inner sep=0.75pt]    {$ \begin{array}{l}
z( \theta )\\
\end{array}$};
\draw (267,185.9) node [anchor=north west][inner sep=0.75pt]    {$\vec{f}( \theta )$};
\draw (118,58) node [anchor=north west][inner sep=0.75pt]   [align=left] {Encoder};
\draw (61,58.4) node [anchor=north west][inner sep=0.75pt]    {$p( \theta )$};
\draw (187,97.5) node [anchor=north west][inner sep=0.75pt]    {$ \begin{array}{l}
z( \theta )\\
\end{array}$};
\draw (262,59.9) node [anchor=north west][inner sep=0.75pt]    {$f( \theta )$};

\end{tikzpicture}
\caption{Top: Learning the weights for a linear combination of features that sums to an approximation for the true fitness function $f(\theta)$. These weights can then be used to predict fitness from measures $m(\theta) = z(\theta)$, or even straight from the phenotype $p(\theta)$. Bottom: De-aggregation of last layer of learned fitness model to result in a set of sub-objectives $\vec{f}(\theta)$ that sum to an approximation of the ground truth fitness $f(\theta)$.}\label{fig:fitness-model}
\end{figure}

Figure~\ref{fig:fitness-model} gives an overview of the training procedure of the fitness prediction model. First, after having trained the autoencoder to a satisfactory level, we freeze the encoder weights, and add a final layer that leads to a single fitness value. We can update the weights of this last layer (through back-propagation, or any other linear regression technique), to accurately predict the fitness value. Then, once the weights are determined, we can \emph{de-aggregate} them into a vector of sub-fitness values. These values correspond well to sub-objectives for multi-objective optimization algorithms.

\subsection{Bringing It All Together}

Given a genotype $g(\theta)$, we can create a phenotype $p(\theta)$ and a gound truth fitness value $f(\theta)$. This is domain specific but does not depend on which optimization algorithm being used. Examples of the phenotype could be the trajectory of a reinforcement learning agent, raw sensory information from a robot, or any other characterization of how the genotype behaves after translation. Then, we train a dimensionality reduction model (such as an autoencoder) to learn a lower dimensional representation of the phenotype $z(\theta)$.

\paragraph{Quality Diversity} Using the lower dimensional representation, we can assign the measures to simply be the values of the latent variables (i.e. $m(\theta) = z(\theta)$). Then, we can assign the fitness value to be the ground truth fitness $f(\theta)$. This is very similar to the procedure done by \citet{cully_autonomous_2019} to automatically discover the measure functions.

\paragraph{Multi-objective Optimization} We can then augment our measure function to include a final layer that predicts the ground truth fitness. Once this model is trained to convergence, we simply de-aggregate the final node, resulting in an element-wise multiplication of the measures and their corresponding weights. These values $\vec{f}(\theta)$ should sum approximately to the ground truth fitness $f(\theta)$. These fitness functions can then be used as the multiple objectives for any MOO scheme.

\paragraph{Fair Comparison?}
What metric do we use to evaluate how these methods are performing? Since the domain is one that ultimately does have a single objective, we should measure the performance of the algorithms by how well they solve this objective. Including a comparison about diversity would not be yield a fair performance comparison as only QD methods explicitly optimize for diversity. So, in the context of solving hard exploration problems, we should be measuring how well each algorithm solves these problems. The authors suggest simply using the ground truth fitness function of the individuals produced through both optimization paradigms.

\subsection{Practical Details}
In practice, using the scheme we presented requires some choices to be made by the user. In this section, we discuss such details and considerations that experimenters attempting to use this comparison scheme should be aware of.

\paragraph{Training the Models}
In order to train the models, one needs to amass a significant amount of training data. \citet{cully_autonomous_2019} solves this issue by using the archive of individuals produced through a run of MAP-elites to train the dimensionality reduction model. As we are attempting to form a comparison scheme between two classes of algorithms, it is important that the model used does not vary much between these two systems. In order to this, the authors see 2 options:

\begin{enumerate}
    \item Pre-trained: Generate the training data from a series of prior evolutionary runs (or perhaps exhaustively enumerating all of genotype space, if tractable), and train the models using that. Use the entire dataset to train the VAE, and then do a second pass to learn the weights of the linear layer given the ground truth fitness as a label. 
    \item Incremental: Train the models with some random phenotypes from randomly sampling from genotype space, and fine-tune the models \emph{differently} for each system, using the phenotypes cached throughout their respective evolutionary runs and the ground truth fitness.
\end{enumerate}

In order to facilitate a fair comparison, the authors recommend method (1). This is because method (2) results in two different models being used to extract the information from phenotype used for selection. This could result in the lack of clarity regarding each system's performance: was it due to actually solving the domain better, or due to generating better training data for the models to later perform better using? 

\paragraph{Dimensionality of the latent layer}

Another practical consideration to be made is the extent to which we shrink dimensions using the dimensionality reduction technique. If the bottleneck layer is too small, it could result in poor reconstruction capability. If the bottleneck layer is too large, it could prohibit effective learning by the optimization algorithms. This is a domain specific consideration, but could be addressed through a series of hyperparameter optimization runs. For example, one could run AURORA \citep{cully_autonomous_2019} using various latent layer sizes, and take the latent layer size that results in the highest QD-Score. As this is simply a benchmark, the cost of training the model can be amortized over many benchmarking tasks.

\section{Conclusion and Future Work}
We have presented a method to compare quality diversity techniques to objective based techniques for a variety of hard exploration or deceptive problems. The key insight is that the comparison to single objective optimization is not a fair comparison due to the aggregation that occurs when we aggregate an individual's performance to be a single number. Each individual solves different sub-objectives to different extents. Optimizing for different combinations of all these sub-objectives could correspondingly also extrinsically optimize for diversity without needing to explicitly be tasked to. This means we can compare the quality of solutions that are created by explicitly optimizing for diversity to directly optimizing for the sub-objectives in their ability to use extrinsic diversity in their favor.

We present a dimensionality reduction technique based on an autoencoder to automatically learn the measure functions from a given phenotype. Then, fitness is approximated as a weighted sum of these measure values. Finally, we remove the summation operation, and set the sub-objectives to simply be an element-wise multiplication of the measure and its weight (i.e. positive weighted measures are increased, negatives decreased). We can perform multi-objective optimization on the sub-objectives, and quality diversity straight on the measures (with access to the ground truth fitness).

Future work in improving this benchmark should address the limitation that the sub-objectives might not be a function of the measures. This could be addressed by training a new fitness prediction architecture that does not re-use weights from the measure predicting autoencoder. Simply de-aggregating the penultimate layer on this model would similarly allow for multiple sub-objectives that are learned straight from the phenotype.

\begin{acks}
The authors acknowledge Antoine Cully, Bill Tozier, Edward Pantridge, Li Ding, Maxime Allard, Nic McPhee, Thomas Helmuth, and the members of the PUSH Lab at Amherst College for discussions and inspiration that helped shape this work. Furthermore, the authors would like to thank the anonymous reviewers for their careful reading and insightful comments and discussion.

This material is based upon work supported by the National Science Foundation under Grant No. 2117377. Any opinions, findings, and conclusions or recommendations expressed in this publication are those of the authors and do not necessarily reflect the views of the National Science Foundation.

\end{acks}

\bibliographystyle{ACM-Reference-Format}
\bibliography{sample-base}

\end{document}